\documentclass[smallextended]{svjour3}       
\smartqed  
\usepackage{graphicx}
\usepackage{enumerate}
\usepackage{amsmath}

\newcommand{\cf}[0]{{\mathcal F}}
\newcommand{\xcf}[0]{x_{\cf}}
\newcommand{\cg}[0]{{\mathcal G}}
\newcommand{\xcg}[0]{x_{\cg}}
\newcommand{\ch}[0]{{\mathcal H}}
\newcommand{\xch}[0]{x_{\ch}}
\newcommand{\enc}[0]{\Leftrightarrow_{enc}}
\newcommand{\la}[0]{\rightarrow}
\newcommand{\uc}[0]{{\rm{ ~unit ~ clause}}}
\newcommand{\ucs}[0]{{\rm{ ~unit ~ clauses}}}
\newcommand{\bcs}[0]{{\rm{ ~binary ~ clauses}}}
\newcommand{\tcs}[0]{{\rm{ ~ternary ~ clauses}}}
\newcommand{\ec}[0]{{\rm{ ~empty ~ clause}}}
\newcommand{\nin}[0]{\not \in}
\newcommand{\false}[0]{false}

\newcommand{\fa}[0]{\forall}
\newcommand{\lra}[0]{\leftrightarrow}

\newcommand{\cfi}[1]{{\mathcal F}_{#1}}
\newcommand{\xcfi}[1]{x_{\cfi{#1}}}

\begin{document}

\title{Set Constraint Model and Automated Encoding into SAT: Application to the Social Golfer Problem
}

\titlerunning{Set Constraint Model and Encoding for SAT}        

\author{Fr\'ed\'eric Lardeux   \and
             Eric Monfroy \and
             Broderick Crawford \and
             Ricardo Soto 
}


\institute{F. Lardeux \at
              Universit\'e d'Angers, France \\
              \email{Frederic.Lardeux@univ-angers.fr}           
           \and
           E. Monfroy \at
           LINA, UMR CNRS 6241, Universit\'e de Nantes, France\\
              \email{Eric.Monfroy@univ-nantes.fr}  
         \and
         B. Crawford \at
         Pontificia Universidad Cat\'olica de Valparaiso, Valparaiso 2362807, Chile\\
        and
        Universidad Finis Terrae, Santiago 7500000, Chile\\
       \email{broderick.crawford@ucv.cl}
        \and
         R. Soto \at
        Pontificia Universidad Cat\'olica de Valparaiso, Valparaiso 2362807, Chile\\
        and
        Universidad Autónoma de Chile, Santiago 7500000, Chile\\
        \email{ricardo.soto@ucv.cl}
}

\date{Received: date / Accepted: date}

\maketitle

\begin{abstract}
On the one hand, Constraint Satisfaction Problems allow one to declaratively model problems. On the other hand, propositional satisfiability problem (SAT) solvers can handle huge SAT instances. We thus present a technique to declaratively model set constraint problems and to encode them automatically into SAT instances. We apply our technique to the Social Golfer Problem and we also use it to break symmetries of the problem. 

Our technique is simpler, more declarative, and less error-prone than direct and improved hand modeling. The SAT instances that we automatically generate contain less clauses than improved hand-written instances such as in~\cite{TriskaMusliu2012}, and with unit propagation they also contain less variables. Moreover, they are well-suited for SAT solvers and they are solved faster as shown when solving difficult instances of the Social Golfer Problem.
\end{abstract}

\section{Introduction}

Most of combinatorial problems can be formulated as Constraint Satisfaction Problems (CSP)~\cite{handbookCP}. A CSP is defined by some variables (generally over finite domains) and constraints between these variables. Solving a CSP consists in finding assignments of the variables that satisfy the constraints. One of the main strength of CSP is declarativity: variables can be of various types (finite domains, floating point numbers, intervals, sets, \ldots) and constraints as well (linear arithmetic constraints, set constraints, non linear constraints, Boolean constraints, symbolic constraints, \ldots). Moreover, the so-called global constraints not only improve solving efficiency but also declarativity: they propose new constructs and relations such as \textit{alldifferent} (to enforce that all the variables of a list have different values), \textit{cumulative} (to schedule tasks sharing resources), \ldots

On the other hand, the propositional satisfiability problem (SAT) \cite{garey79} is restricted (in terms of declarativity) to Boolean variables and propositional formulae. However, SAT solvers can now handle huge SAT instances (millions of variables). It is thus attractive to~1) encode CSPs into SAT (e.g., \cite{bacchusCP2007,bessiereSAT2003}) in order to benefit from the declarativity of CSP and the power of SAT, or~2) introduce more declarativity into SAT, for example with global constraints (e.g., alldifferent~\cite{micai2009}, cardinality~\cite{Bailleux03}).

\medskip

In this paper we are concerned with the transformation of set constraints into SAT instances: we often refer to this transformation as "encoding". Various systems of set constraints (either specialized systems~\cite{CLPS}, libraries for constraint programming systems such as~\cite{ConjuntoILPS94}, or the set constraint library of CHOCO~\cite{choco}) have been designed for solving problems such as prototyping combinatorial problems, axiomatization of set theory, analysis of programs,\ldots They have shown that some problems can easily be modeled with set constraints.

\medskip

Coding set constraints directly into SAT is a tedious tasks (see for example~\cite{TriskaMusliu2012} or~\cite{GentLynce2005}). Moreover, when one wants to optimize its model in terms of variables and clauses this quickly leads to very complicated and unreadable models in which errors can easily appear. Thus, our approach is based on an automated encoding of set constraints into SAT instances. To this end, we define some encoding rules ($\Leftrightarrow_{enc}$) that encode set constraints (such as intersection, union, membership, cardinal of sets) into the corresponding SAT clauses and variables. The advantage is that the modeling language (i.e., standard set constraints) is declarative, simple, and readable. 
We have tried this technique on various problems, and the SAT instances which are automatically generated have a complexity similar to the complexity of improved hand-written SAT formulations, and their solving with a SAT solver (in our case Minisat) is efficient.

\medskip

We illustrate our approach with the Social Golfer Problem (problem number 10 of the CSPLib~\cite{csplib}). The problem is the following: $q$ golfers play every weeks during $w$ weeks split in $g$ groups of $p$ golfers ($q=p.g$). How to schedule the play of these golfers such that no golfer plays in the same group as any other golfer more than once. An instance of the problem is then given by a triple $g-p-w$. Various instances of the Social Golfer Problem are still open, and the problem is attractive since it is  related to problems  such as encryption and covering problems. Compared to direct encodings (such as the one of~\cite{TriskaMusliu2012}), the instances we generate are smaller (less clauses), and also contain less variables using unit propagation. The introduction of symmetry breaking is simplified with our technique and can be done by adding constraints to the initial model or by refining the initial model.
Using Minisat~\cite{minisat03}, our automatically generated instances (with or without symmetry breaking) are solved  faster than the ones of~\cite{TriskaMusliu2012}.

\medskip

We can compare our work with works of different types, first of all with SAT encoding techniques such as~\cite{bacchusCP2007} and \cite{bessiereSAT2003}. These works make a relation between CSP solving and SAT solving in terms of properties such as consistencies for finite domain variables and constraints. In this article, we are concerned with a different type of constraints, i.e., set constraints.

Concerning applications, i.e., the Social Golfer Problem, the closest work is~\cite{TriskaMusliu2012} which is a revision and improvement of~\cite{GentLynce2005}. Whereas these works are hand-written modeling of the Social Golfer Problem directly in SAT, we are concerned with a higher-level model language which is automatically transformed into SAT instances. \cite{TriskaMusliu2012} also 
proposes various symmetry breaking techniques to improve the model; some of these symmetries naturally disappear using our set constraint model (for example, we do not have the permutations due to numbering of groups within a week). The remaining symmetry breakings can easily be introduced in our model, by adding constraints or by refining the initial model.

In~\cite{sgpVH2006}, the Social Golfer Problem is modeled with a combination of set constraints and arithmetic constraints. However, this model is not directly used but it is transformed into CSP before being solved by mimetic algorithms.

Finally, our approach is similar to~\cite{micai2009} in which alldifferent global constraints and overlapping alldifferent constraints are handled declaratively before being encoded automatically in SAT using rewrite rules. Note also that we use the work of~\cite{Bailleux03} about the \textit{cardinality} global constraint in order to perform the encoding of set cardinality.

\medskip

In the next section (Section~\ref{sec:encoding}), we present our set constraint language and the rule-based system for encoding set constraints into SAT; we consider standard set constraints. To get a comparison basis, we then (Section~\ref{models}) give a direct SAT model of the Social Golfer Problem, and some variants of this model.  We then present how to model the Social Golfer Problem with set constraints, and show the interest of our system in terms of declarativity. 
In Section~\ref{sb}, we show how to introduce symmetry breaking techniques ( that can be found in the literature) with our set constraint language:  by adding new constraints or by refining the initial model. 
In Section~\ref{sec:compmod}, we compare various SAT instances, either hand-written our automatically generated with our encoding rule: this analysis is made with respect to instance structures  (e.g., number of variables and clauses). In the next section, we compare the solving time of these instances.
Section~\ref{discussion}, discusses various points related to our technique: structure of instances, usefullness of unit propagation, difference with work about set constraints in constraint programming, \ldots
We finally conclude in Section~\ref{conclusion}.

\section{Set Constraint Encoding}
\label{sec:encoding}

We present here the encoding of usual (CSP) set constraints (such as $\in$, $\cup$, $\cap$, \ldots) into SAT clauses.
More constraints could be defined, but they can be deduced from these basic constraints.

\subsection{Universe and Supports}

We consider two notions: \textit{universe} and \textit{support}. Unformally, the universe is the set of all elements that are considered in a model of a given problem while the support ${\cal F}$ of a set $F$ appearing in this model is a set of possible elements of $F$  (i.e., ${\cal F}$ is a superset of $F$).

\begin{definition}
Let $P$ be a problem, and $M$ be a model of $P$ in ${\cal L}$, i.e., a description of $P$ from the natural language to the language of constraints ${\cal L}$. 
\begin{itemize}
\item The universe ${\cal U}$ of $M$ is a finite set of constants.
\item The support of the set $F$ of the model $M$ is a subset of the universe ${\cal U}$; we denote it by ${\cal F}$.  ${\cal F}$ represents the elements of ${\cal U}$ that can possibly be elements of $F$:
\[
F \subseteq {\cal F} \subseteq {\cal U} {\textrm ~~~and~~~} F \in {\cal P}({\cal F})
\]
where ${\cal P}({\cal F})=\{A | A \subseteq {\cal F} \}$ is the power set of ${\cal F}$. We say that $F$ is over  ${\cal F}$.
\end{itemize}

\end{definition}

Note that each element of~${\cal U} \setminus {\cal F}$ cannot be  element of $F$.
In the following, we denote sets by uppercase letters (e.g., $F$) and their supports by calligraphic uppercase letters (e.g., ${\cal F}$). When there is no confusion of model, we shorten "the set $F$ of the model $M$" to "the set $F$".

\medskip

Consider a model $M$ with a universe ${\cal U}$, and a set $F$ over ${\cal F}$. For each element $x$ of ${\cal F}$, we consider a Boolean variable $x_{\cal F}$ which is true if $x \in F$ and false otherwise. 
We call the set of such variables the support variables for $F$ in ${\cal F}$.

\begin{example}
Let ${\cal U}=\{x, y, z, t\}$ be the universe of a model $M$, and ${\cal F}=\{x,y, t\}$ be the support of a set $F$ of $M$. 
Then, we have 3 Boolean variables $x_{\cal F}$, $y_{\cal F}$, and $t_{\cal F}$ corresponding respectively to $x$, $y$, and $z$ to represent $F$. By definition, $z \not \in F$ and there is no $z_{\cal F}$ variable; and $x,y, t$ can possibly be  in $F$.
Consider now that $F=\{x,y\}$. Then, $x_{\cal F}=true$, $y_{\cal F}=true$, and $t_{\cal F}=false$
\end{example}
\bigskip

In the following, we write $x_{\cal F}$ for $x_{\cal F}=true$ and $\neg x_{\cal F}$ for $x_{\cal F}=false$.

\subsection{The  $\Leftrightarrow_{enc}$ Encoding Rule}

We can now define the encoding of the various CSP set constraints into SAT. 
In the following, we consider three sets $F$, $G$, and $H$ respectively defined on the supports $\cf$, $\cg$ and $\ch$ of the universe ${\cal U}$, and for each $x \in {\cal U}$ the various Boolean variables $x_{\cal F}$ , $x_{\cal G}$, and $x_{\cal H}$ as defined before. $|G|$ denotes the cardinality of the set $G$.

 Note that we do not force the supports to be minimal: for example,
for the equality constraint $F=G$, the sets $\cf \setminus \cg$ and $\cg \setminus \cf$ can be non empty whereas $F \setminus G$ and $G \setminus F$ must be empty. We thus consider these cases in the  $\Leftrightarrow_{enc}$ encoding rule. Allowing the supports to be non minimal eases the modeling process: indeed, one does not have to compute the minimal support and can use a superset of it or the universe. This is practical when sets are built from many other sets using numerous set constraints. Note also that using the minimal supports reduces the size of the generated SAT instances.

The  encoding rule is noted $\Leftrightarrow_{enc}$. The clauses that are generated by this rule are of the form $\fa x \in \cf,~ \phi(\xcf)$ which denotes the $|\cf|$ formulae $\phi(\xcf)$ built for each element  $x$ of the support $\cf$ of $F$ ($x$ refers to the element of the universe/support, and $\xcf$ to  the variable representing $x$ for the set $F$). For the membership constraint, the rule is not quantified; for multi-intersection and multi-union, an additional universal quantifier over $i$  is used to denote a set of encoding rules, each rule being related to one of the sets $\cfi{i}$.

In the following, we propose several set constraint encodings with: first the set constraint, then its encoding in SAT, and finally, the number of clauses generated.

\subsection{Membership Constraint}
This constraint enforces the membership of an element $x$ to a set $F$:
\begin{itemize}
\item if $x \in \cf$ ($x$ is in the support of $F$), then the corresponding support variable must be true, i.e., $ \xcf$.
\item  if $x \not \in \cf$ ($x$ is not in the support of $F$), then the constraint $x \in F$ must generate a failure since the problem does not have any solution.
\end{itemize}

\[
\begin{array}{l l l}
x \in F & 
\enc &
\left \lbrace 
\begin{array}{l@{~~~~} l}
x \in \cf,~ \xcf & 1 \uc \\
x \nin \cf,~ \false & 1 \ec \\
\end{array}
\right.
\end{array}
\]

The constraint $x \not \in F$ can be similarly defined.

\subsection{Set Equality Constraint}
Two sets $G$ and $F$ are equal if and only if:
\begin{itemize}
\item for the elements of $\cf \cap \cg$:  the support variables of $G$ have the same values as the support variables of $F$;
\item for the elements of $\cf  \setminus \cg$: the support variables of $F$ must be false. Indeed, an element of the universe which is not in the support of a set is not part of this set; thus, an element of $\cf  \setminus \cg$ cannot be in $F$.
\item for the elements of $\cg  \setminus \cf$: the support variables of $G$ must be false.
\end{itemize}

\[
\begin{array}{l l l}
F = G& 
\enc &
\left \lbrace 
\begin{array}{l@{~~~~} l}
\fa x \in \cf \cap \cg,~ \xcf \lra \xcg &   2.|\cf \cap \cg| \bcs\\
\fa x \in \cf \setminus \cg,~ \neg \xcf &  |\cf \setminus \cg| \ucs\\
\fa x \in \cg \setminus \cf,~ \neg \xcg &  |\cg \setminus \cf| \ucs\\
\end{array}
\right.
\end{array}
\]

The constraint $F \neq G$ can be similarly defined by considering the negation of the conjunction of formulae of the previous encoding.

\subsection{Intersection Constraint}
Let $H$ be the intersection of two sets $G$ and $F$:
\begin{itemize}
\item for the elements of $\cf \cap \cg \cap \ch$: a support variable of $H$ is true if and only if this variable is in $F$ and $G$;
\item for the elements of  $ (\cf \cap \cg) \setminus \ch$: since such an element cannot be in $H$, it must not be in $F$ and $G$;
\item for the elements of  $ \ch \setminus (\cf \cap \cg)$: a support variable of $H$ which is not in the support of $F$ and $G$ cannot be true
\end{itemize}
\[
\begin{array}{c}
F \cap G = H  \\
 \enc \\
\left \lbrace 
\begin{array}{l@{~~~~} l}
\fa x \in \cf \cap \cg \cap \ch,~ \xcf \wedge \xcg \lra \xch & 
\begin{array}{@{} l}
|\cf \cap \cg \cap \ch| \tcs \\
+ 2. |\cf \cap \cg \cap \ch| \bcs
\end{array}\\
\fa x \in (\cf \cap \cg) \setminus \ch,~ \neg \xcf \vee \neg \xcg&  |(\cf \cap \cg) \setminus \ch| \bcs\\
\fa x \in \ch \setminus (\cf \cap \cg), ~\neg \xch&  |\ch \setminus (\cf \cap \cg)| \ucs
\end{array}
\right.
\end{array}
\]

Note that if $H=\emptyset$ (e.g., we want to force the intersection to be empty), then the encoding  can be simplified into $\forall x \in U,   \neg  x_{F}\vee \neg x_{G}$, and thus, reduce its size to 
 $|U|$ clauses.

\subsection{Union Constraint}
More cases are to be considered for this constraints:
\begin{itemize}
\item  for the elements of $\cf \cap \cg \cap \ch$: a support variable of $H$ is true if and only if this variable is in $F$ or in $G$; this is the trivial case;
\item  for the elements of $(\cf \cap \ch) \setminus \cg$: this case is a reduction of the previous one but it is however equivalent; since such an element $x$ is not in the support of $G$ then $\xcg$ does not exist, and $x$ is in $H$ if and only if it is in $F$; note that the generated clauses are exactly the same removing $\xcg$;
\item for the elements of $(\cg \cap \ch) \setminus \cf$: this is the symmetrical case for $G$;
\item for the elements of $\ch \setminus (\cf \cup \cg)$: the support variables of $H$ that are not in $F$ or in $G$  must be false;
\item for the elements of $\cf  \setminus \ch$: elements of the support of $F$ that are not in the support of $H$ cannot be in $F$;
\item for the elements of $\cg  \setminus \ch$:  symmetrical case for $G$.
\end{itemize}
\[
\begin{array}{c}
F \cup G = H \\
\enc \\
\left \lbrace 
\begin{array}{l@{~~~~} l}
\fa x \in \cf \cap \cg \cap \ch,~ \xcf \vee \xcg \lra \xch & 
\begin{array}{@{} l}
|\cf \cap \cg \cap \ch| \tcs \\
+ 2. |\cf \cap \cg \cap \ch| \bcs
\end{array}\\
\fa x \in (\cf \cap \ch) \setminus \cg,~ \xcf \lra \xch& 2.|(\cf \cap \ch) \setminus \cg| \bcs\\
\fa x \in (\cg \cap \ch) \setminus \cf,~ \xcg \lra \xch& 2.|(\cg \cap \ch) \setminus \cf| \bcs\\
\fa x \in \ch \setminus (\cf \cup \cg),~ \neg \xch&|\ch \setminus (\cf \cup \cg)| \ucs\\
\fa x \in \cf  \setminus \ch,~ \neg \xcf &|\cf \setminus \ch| \ucs\\
\fa x \in \cg \setminus \ch,~ \neg \xcg&|\cg \setminus \ch| \ucs\\
\end{array}
\right.
\end{array}
\]

\subsection{Inclusion Constraint}
\begin{itemize}
\item  for the elements of $\cf \cap \cg $: such an element is in $G$ if it is in $F$,
\item  for the elements of $\cf \setminus \cg$: since these elements cannot be in $G$, they cannot be in $F$;
\end{itemize}
\[
\begin{array}{l l l}
F \subseteq G& 
\enc &
\left \lbrace 
\begin{array}{l @{~~~~}l}
\fa x \in \cf \cap \cg,~ \xcf \rightarrow  \xcg & |\cf \cap \cg| \bcs\\
\fa x \in \cf \setminus \cg,~ \neg \xcf& |\cf \setminus \cg| \ucs
\end{array}
\right.
\end{array}
\]

\subsection{Difference Constraint}

\begin{itemize}
\item  for the elements of $\cf \cap \cg \cap \ch$: such elements are in $H$ if and only if they are in $F$ and not in $G$;
\item  for the elements of $\cf \setminus (\cg \cup \ch)$:  such elements cannot be in $F$;
\item  for the elements of $\ch \setminus \cf$:  such elements cannot be in $H$;
\item  for the elements of $(\cf \cap  \ch)\setminus \cg$: such elements are in $H$ if and only if they are in $F$;
\item  for the elements of $(\cf \cap \cg) \setminus \ch$: since such elements cannot be in $H$, if they are in $F$ they also must be in $G$;
\end{itemize}
\[
\begin{array}{c}
H = F \setminus G\\
\enc \\
\left \lbrace 
\begin{array}{l@{~~~~} l}
\fa x \in \cf \cap \cg \cap \ch,~ \xcf \wedge \neg \xcg \lra \xch &
\begin{array}{@{} l}
 |\cf \cap \cg \cap \ch| \tcs\\
+ 2. |\cf \cap \cg \cap \ch| \bcs
\end{array}\\

\fa x \in \cf \setminus (\cg \cup \ch),~ \neg \xcf & |\cf \setminus (\cg \cup \ch)| \tcs\\

\fa x \in \ch \setminus \cf,~ \neg \xch & |\ch \setminus \cf| \ucs\\

\fa x \in (\cf \cap  \ch)\setminus \cg,~ \xcf \lra \xch & 2.|(\cf \cap  \ch)\setminus \cg| \bcs\\

\fa x \in (\cf \cap \cg) \setminus \ch,~ \xcf \la \xcg & |(\cf \cap \cg) \setminus \ch| \bcs\\
\end{array}
\right.
\end{array}
\]

\subsection{Multi-union Constraint}

The multi-union constraint $H =\bigcup_{i=1}^{n} F_{i}$ is  equivalent to the $n$ constraints expressed as $H=F_1 \cup (F_2 \cup ( \ldots (F_{n-1} \cup F_n) \ldots  )$. It is not only a short-hand, but it also significantly reduces the number of generated clauses. Indeed, elements of $\bigcap_{i=1}^{n} \cfi{i}$ are considered once in the multi-union constraint whereas it is considered $n$ times in the corresponding $n$ union constraints.
We do not detail the encoding since this is an extension of the union constraint.
In the next formulae, the set $\{1,\ldots,n\}$ is noted $N$.
\[
\begin{array}{c}
H =\bigcup_{i=1}^{n} F_{i} \\
\enc \\
\left \lbrace 
\begin{array}{l@{~~} l}
\begin{array}{l}
\fa I,J \in {\cal P}(N), I\neq \emptyset, I\cup J=N,\\
\fa x \in \ch \cap  (\bigcap_{i\in I} \cfi{i})\setminus (\bigcup_{j\in J} \cfi{j}), \bigvee_{i\in I} \xcfi{i}\lra \xch
\end{array}
&
(I)
\\
&\\
\fa x \in \ch \setminus  (\bigcup_{i=1}^n \cfi{i}),~ \neg  \xch
&
(II)
\\
\fa i \in [1..n] ,~\fa x \in \cfi{i} \setminus \ch ,~ \neg  \xcfi{i}
&
(III)
\\
\end{array}
\right.
\end{array}
\]

\begin{itemize}
\item[] $(I)$ generates  
\[
\begin{array}{l}
\sum\limits_{{I,J \in {\cal P}(N),\atop{ I\neq \emptyset, \atop{I\cup J=N}}}}\big( | \ch \cap  (\bigcap_{i\in I} \cfi{i})\setminus (\bigcup_{j\in J} \cfi{j})|.(|I|+1)\big)~\bcs \\
\textrm{and}\\ 
\sum\limits_{{I,J \in {\cal P}(N),\atop{ I\neq \emptyset, \atop{I\cup J=N}}}} \big( | \ch \cap  (\bigcap_{i\in I} \cfi{i})\setminus (\bigcup_{j\in J} \cfi{j})|\big)~~ (|I|+1)\textrm{-ary clauses}
\end{array}
\]
\item[] $(II)$ generates  $|\ch \setminus (\bigcup_{i=1}^n \cfi{i})|  \ucs$ 
\item[] $(III)$ generates  $ \sum_{i=1}^{n} |(\cfi{i} \setminus \ch|  \ucs$ 
\end{itemize}

Note also that in our implementation that generates SAT instances, the result of an union must be stored in a set: thus, $H =\bigcup_{i=1}^{n} F_{i}$ is equivalent to $H=F_1 \cup H_1$, $H_1 =F_2 \cup H_2$, \ldots $H_{n-1}=F_{n-1} \cup F_n$. The multi-union constraint thus also significantly reduce the number of variables (variables necessary for the intermediate sets $H_i$).

\subsection{Multi-intersection Constraint}

Similarly, we define the multi-intersection constraints. As for the multi-union, the advantage is the gain of clauses, and of variables in our implementation of the encoding.

\[
\begin{array}{lll}
H =\bigcap_{i=1}^{n} F_{i}  &
\enc &
\left \lbrace 
\begin{array}{l@{~~~~} l}
\fa x \in \ch \cap (\bigcap_{i=1}^{n} \cfi{i}),~ \bigwedge_{i=1}^{n} \xcfi{i} \lra \xch
&
(I)
\\
\fa x \in  \bigcap_{i=1}^{n} \cfi{i} \setminus \ch,~ \bigvee_{i=1}^{n} (\neg \xcfi{i}) 
& 
(II)
\\
\fa x \in \ch \setminus (\bigcap_{i=1}^{n} \cfi{i}),~   \neg \xch
&
(III)
\\
\end{array}
\right.
\end{array}
\]

\begin{itemize}
\item[] $(I)$ generates  $2. |\ch \cap (\bigcap_{i=1}^{n} \cfi{i})| ~(n+1)\textrm{-ary clauses}$ 
\item[] $(II)$ generates  $|\bigcap_{i=1}^{n} \cfi{i} \setminus \ch| ~n\textrm{-ary clauses}$ 
\item[] $(III)$ generates  $|\ch \setminus (\bigcap_{i=1}^{n} \cfi{i})| \ucs$ 
\end{itemize}

\subsection{Cardinality Constraint}
This constraint is interesting to enforce the size of a set, or to compute the size of a set.
We denote by $k=|G|$ the cardinality constraint linking the cardinal of $G$ to the finite domain number (or variable) $k$. This constraint has been studied for the encoding of global constraints, see for example \cite{Bailleux03}.
 
 The very intuitive encoding of this constraint is quite simple. If we have a support ${\cal G}$ of size $n$ and we want to obtain a set $G$ of $k$ elements ($k\leq n$) we have to verify that:
 \begin{itemize}
 \item All the sets of $k+1$ variables have at least one false variable.
 \item All the sets of $n-k+1$ variables have at most one true variable.
\end{itemize}
\[
|G|=k~~ \Leftrightarrow_{enc}\]
\[ \forall \{x_1,...,x_{k+1}\} \subseteq {\cal V}, \bigvee_i  \neg x_i, \forall \{x_1,...,x_{n-k+1}\} \subseteq {\cal V}, \bigvee_i  x_i
\]

The weakness of this encoding is the number of generated clauses:
\[\frac{n!}{(k+1)!+(n-k-1)!} + \frac{n!}{(k-1)!+(n-k+1)!}\]

A more efficient encoding (but less intuitive) for this constraint is the use of the unary representation of integers (an integer $k \in[0..n]$ is represented by 1 $k$ times followed by 0 $n-k$ times). This encoding is presented in \cite{Bailleux03} with two main components: the \textit{totalizer} and the \textit{comparator}. Note that we have chosen this encoding for the unit clauses it generates (see Section~\ref{sec:up}).
 
 The totalizer corresponds to a balanced binary tree structure. It is used to associate an auxiliary variable (output variable) for each variable of the cardinality constraint (input variable) and to sort these new variables such that the true variables are placed before the false variables. Internal variables used to linked input and output variables are called  linking variables. The main property of the binary tree is that each non-leaf node corresponds to the union of the two children. The leaves are the input variables and the seed is the set of the output variables. Each node $N$  has two child nodes $C^1$ and $C^2$ that are sets of Boolean variables. We denote $C^1_\alpha$ the $\alpha$-th variable of the set $C^1$. 

The totalizer is encoded by generating for each node the next clauses: 
 \[
\bigwedge_{{0\leq \alpha \leq|C^1| \atop
0\leq \beta \leq|C^2|} \atop
{0\leq \gamma\leq|N| \atop
\alpha+\beta= \gamma}}  (\neg C^1_\alpha \vee \neg C^2_\beta \vee N_ \gamma) \wedge  (C^1_{\alpha+1} \vee C^2_{\beta+1} \vee \neg N_{\gamma+1})
\]
with 
\begin{itemize}
\item $C^1_0=C^2_0=N_0=1$
\item $C^1_{|C^1|+1}=C^2_{|C^2|+1}=N_{|N|+1}=0$
\end{itemize}

The comparator enforces the cardinal $k$ of the set simply by assigning the true value to the first $k$ output variables (noted $s_i$) of the totalizer. Its encoding is very simple:
 \[
\bigwedge_{1\leq i \leq k} s_i \bigwedge_{k+1\leq j \leq n} \neg s_j
\]

In total, if $G$ is over the support ${\cal G}$ of size $n$, then the set constraint $|G|=k$ generates:
\begin{itemize}
\item  \textbf{$n+ \sum_{i=1}^{n}{2u_i^{n}(\lfloor \frac{u_i^{n}}{2}\rfloor+1)(\lceil \frac{u_i^{n}}{2}\rceil+1)-(\frac{u_i^{n}}{2}+1)} $} clauses 
\item \textbf{$\sum_{i=1}^{n}{u_i^n}$} variables.
\end{itemize}
with $u_n^n=1$,$u_1^n=n$ and  $u_i^n=u_{2i-1}^n+2u_{2i}^n+u_{2i+1}^n$.

\section{Models for the Social Golfer Problem}
\label{models}

In this section we describe various SAT related models for the Social Golfer Problem.

\subsection{Direct Encoding}
\label{sec:directmodel}

In order to present (and then compare) a SAT model for the Social Golfer Problem which does not use set constraints, we give here a model, similar to the one of~\cite{TriskaMusliu2012}  (which was already a revision of~\cite{GentLynce2005}) without auxiliary variables.

The Boolean variables to be considered are denoted $g_{q',p',g',w'}$ meaning (when $g_{q',p',g',w'}$ is true) that player $q'$ is the $p'$-th player of the group number $g'$ of week $w'$ with:
\begin{itemize}
\item $p'$ ranging from 1 to $p$, $p$ being the number of players in each group;
\item $g'$ ranging from 1 to $g$, $g$ being the number of groups each week;
\item $q'$ ranging from 1 to $q$, $q=g.p$ being the total number of players;
\item and $w$ ranging from 1 to $w$, $w$ being the number of weeks considered.
\end{itemize} 
With the $q.p.g.w$  variables of type $g_{q',p',g',w'}$, the constraints are:
\begin{itemize}
\item each golfer plays once per week;
\item there is $p$ players in each group;
\item two players never play twice in the same group.
\end{itemize}

\paragraph{Each golfer plays at least once per week}
To enforce that each golfer plays at least once per week, we need the following $g.p.w$ clauses:
\begin{eqnarray}
\bigwedge_{q'=1}^{q}
\bigwedge_{w'=1}^{w}
\bigvee_{p'=1}^{p}
\bigvee_{g'=1}^{g}
g_{q',p',g',w'} \label{atleastonceperweek}
\end{eqnarray}
meaning that for each week $w'$, each player $q'$  is at least the $p'$-th player in one group $g'$.

\paragraph{Each players plays at most once per week}
Enforcing that each players plays at most once per week is done in two steps, first enforcing that each golfer plays at most once per group in each week: on week $w'$, group $g'$, the same player cannot play both on position $p'$ of $g'$ and position $p''$ of $g'$:
\begin{eqnarray}
\bigwedge_{q'=1}^{q}
\bigwedge_{w'=1}^{w}
\bigwedge_{p'=1}^{p}
\bigwedge_{g'=1}^{g}
\bigwedge_{p''=p'+1}^{p}
\neg g_{q',p',g',w'} ~\vee~ \neg g_{q',p'',g',w'}\label{atmostperweek}
\end{eqnarray}
Formula~(\ref{atmostperweek})  consists in $q.w.g.p.(p-1)/2$ clauses.

Then, the following $q.w.p.(p-1).g.(g-1)/4$ clauses ensure than a player does not play in more than a group each week:
\begin{eqnarray}
\bigwedge_{q'=1}^{q}
\bigwedge_{w'=1}^{w}
\bigwedge_{p'=1}^{p}
\bigwedge_{g'=1}^{g}
\bigwedge_{g''=g'+1}^{g}
\bigwedge_{p''=p'+1}^{p}
\neg g_{q',p',g',w'} ~\vee~ \neg g_{q',p'',g'',w'} \label{onlyonegroup}
\end{eqnarray}

\paragraph{Groups are correct}
The same has to be done for groups to ensure that they are correct: one and only one player per position in each group, each week.  There is at least a golfer playing at position $p'$ in the group $g'$ on week $w'$; this gives $w.p.g$ clauses:
\begin{eqnarray}
\bigwedge_{w'=1}^{w}
\bigwedge_{p'=1}^{p}
\bigwedge_{g'=1}^{g}
\bigvee_{q'}^{q}
g_{q',p',g',w'}  \label{oneplayerperposition}
\end{eqnarray}
And at most one golfer plays at position $p'$ in the group $g'$ on week $w'$:
\begin{eqnarray}
\bigwedge_{w'=1}^{w}
\bigwedge_{p'=1}^{p}
\bigwedge_{g'=1}^{g}
\bigwedge_{q'}^{q}
\bigwedge_{q''=q'+1}^{q}
\neg g_{q',p',g',w'} ~\vee~ \neg g_{q'',p',g',w'}\label{atmostoneplayerperposition}
\end{eqnarray}
which results in $q.(q-1).w.p.g/2$ clauses.

\paragraph{The socialization constraint}
The only remaining constraint (named the socialization constraint) states that two players cannot play twice in the same group, i.e., if a player $q'$ plays in the same group $g'$ on the same week $w'$ as player $q''$, and that $q'$ plays in another group $g''$ another week $w''$, then $q''$ cannot play on group $g''$ on week $w''$ at whatever position:
\[
\bigwedge_{w'=1}^{w}
\bigwedge_{g'=1}^{g}
\bigwedge_{w''=w'+1}^{w}
\bigwedge_{g''=1}^{g}
\bigwedge_{q'=1}^{q}
\bigwedge_{p_1=1}^{p}
\bigwedge_{p_1'=1}^{p}
\bigwedge_{q''=q'+1}^{q}
\bigwedge_{p_2=1}^{p}
\bigwedge_{p_2'=1}^{p}
\]
\begin{eqnarray}
g_{q',p_1,g',w'} \wedge g_{q'',p_2,g',w'} 
\wedge g_{q',p_1',g'',w''} \rightarrow \neg  g_{q'',p_2',g'',w''}\label{social}
\end{eqnarray}
Formula~(\ref{social}) is the hard point of the direct model with a complexity of $w.(w-1).g^2.q.(q-1).p^4/4$ clauses.

\paragraph{Complexity of the direct encoding}
The complexity of the direct encoding DE which contains Formulae~(\ref{atleastonceperweek})--(\ref{social}) is thus:
${\cal O}(w^2.g^4.p^6)$
 in terms of clauses with $p^2.g^2.w$ variables.

\subsection{Variants of the Direct Encoding}

\subsubsection{The Ladder matrix structure} 
In \cite{GentLynce2005} a ladder matrix is used: the ladder matrix, which was first presented in~\cite{ladder},  introduces a set of auxiliary variables $g'_{i,k,l} \leftrightarrow \bigvee_{p'=1}^{p} g'_{i,p',k,l}$. Intuitively, these new variables abstract the positions of the players in the group. These new variables together with the characteristics of the ladder matrix are then used to model the socialization constraint. The resulting constraints are a bit less complex than the socialization constraint given above, but the ladder matrix introduces an "intermediate level" in the model which is not so simple to handle and not declarative. Moreover, it also results from this model more variables and more clauses.

\subsubsection{Intermediate variables}
In \cite{TriskaMusliu2012}, $q.g.w$ intermediate variables $g'_{i,k,l}$ are introduced:
\begin{eqnarray}
\fa i \in [1..q], \fa k \in [1..g], \fa l \in [1..w], g'_{i,k,l} \leftrightarrow \bigvee_{p'=1}^{p} g_{i,p',k,l} \label{tmvar}
\end{eqnarray} 
As for the ladder matrix, these variables abstract the positions of players in the groups. These variables simplify the socialization constraint by abstracting positions as follows:
\begin{eqnarray}
\begin{array}{c}
\bigwedge_{w'=1}^{w}
\bigwedge_{g'=1}^{g}
\bigwedge_{w''=w'+1}^{w}
\bigwedge_{g''=1}^{g}
\bigwedge_{q'}^{q}
\bigwedge_{q''=q'+1}^{q} 
\\ 
~~\\
(\neg g'_{q',g',w'} \vee \neg g'_{q'',g',w'})
\vee
(\neg g'_{q',g'',w''} \vee \neg g'_{q'',g'',w''}) 
\end{array}\label{tmc}
\end{eqnarray}
This introduces $q.w.g$ new intermediate variables $g'_{i,k,l}$ and $q.w.g.(p+1)$ clauses in $g'_{i,k,l} \leftrightarrow \bigvee_{p'=1}^{p} g'_{i,p',k,l}$, but this significantly reduces the complexity of the new socialization constraint from $w.(w-1).g^2.q.(q-1).p^4/4$ to $w.(w-1).g^2.q.(q-1)/4$. 

The complexity of the Triska-Musliu encoding~\cite{TriskaMusliu2012} (Formulae~(\ref{atleastonceperweek})--(\ref{atmostoneplayerperposition}), (\ref{tmvar}), and~(\ref{tmc})) is thus ${\cal O}(w^2.g^4.p^2)$
 in terms of clauses. In the following we call this encoding TME.
A more complete analysis in terms of variables and clauses is given in Section~\ref{modelstructure}.

\subsection{SAT Encoding for Set Constraint Model }
We propose a model for the Social Golfer Problem using set constraints in a solver independent way. These constraints are then encoded into SAT using our $\Leftrightarrow_{enc}$ rules.

\subsubsection{Set constraints model}
An instance of the problem is thus given by a triple $g-p-w$:
\begin{itemize}
\item $p$ is the number of players per group;

\item $g$ is the number of groups per week;

\item $w$ is the number of weeks;

\end{itemize}

The universe for this model is the set of players ${\cal P}=\{p_1, \ldots, p_{q}\}$ with $q=g.p$  being the total number of players.
We need the following  $w.g$ set variables to model the  groups $G_{1,1}$,  \ldots, $G_{w,g}$.  The set  $G_{i,j}$ is the group number $j$ of week $i$ and is over the support ${\cal G}_{i,j}={\cal P}$.  Each $G_{i,j}$ will contain $p$ players from ${\cal P}$.
Note that the supports are minimal and cannot be reduced without loosing solutions (or symmetric solutions).
We now give the constraints of the Social Golfer Problem.

\paragraph{$p$ players per group every weeks:}
\begin{eqnarray}
\forall i \in [1..w], \forall j \in [1..g], ~~ |G_{i,j}|=p \label{s1}
\end{eqnarray}

\paragraph{Every golfer plays every weeks:}
\begin{eqnarray}
\forall i \in [1..w]~  \bigcup_{j=1..g} G_{i,j} = {\cal P} \label{s3}
\end{eqnarray}

\paragraph{No golfer plays in two groups the same week:}
\begin{eqnarray}
\forall i \in [1..w]~ \bigcap_{j=1..g} G_{i,j} = \emptyset \label{s2}
\end{eqnarray}
However, Constraints~(\ref{s2})  are not required since they are implied by Constraints~(\ref{s1}) and Constraints~(\ref{s3}).

\paragraph{Two players cannot play twice together in the same group:}

The simplest formulation is:
$\forall p_1, p_2 \in {\cal P}, 
 \forall w_1, w_2  \in [1..w],
\forall g_1, g_2 \in [1..g],$ 
$p_1 \not = p_2 
~\wedge~  (g_1 \not = g_2 \vee w_1 \not = w_2)
 ~\wedge~ p_1 \in G_{g_1,w_1} ~\wedge~ p_2 \in G_{g_1,w_1} ~\wedge~ p_1 \in G_{g_2,w_2}
    ~\rightarrow~  p_2 \not \in  G_{g_2,w_2}$
meaning : if two different golfers play in the same group $g_1$, if $p_1$ plays in another group $g_2$ then $p_2$ cannot play in this group $g_2$. However, due to the permutations $p_1,p_2$, $w_1,w_2$, and $g_1,g_2$, this constraint introduces redundancies that can be removed using the following constraint:
\begin{eqnarray}
\forall w_1, w_2 \in [1..w], p_i, p_j \in {\cal P}, 
g_1, g_2 \in [1..g],  \notag \\
w_1 > w_2 ~\wedge~ i > j ~\wedge~ g_1 \geq g_2  ~\wedge   \label{s4} \\
p_i \in G_{w_1,g_1} ~\wedge~ p_j \in G_{w_1,g_1} ~\wedge~ p_i \in G_{w_2,g_2}
~\rightarrow~ 
p_j \not \in G_{w_2,g_2}\notag
\end{eqnarray}

Another formulation of these constraints can be given using the cardinality constraints:
\begin{eqnarray}
\forall w_1, w_2 \in [1..w], 
g_1, g_2 \in [1..g],  \notag \\
w_1 > w_2 ~\wedge~ 
 g_1 \geq g_2  ~\wedge   \label{s5}\\ 
|G_{w_1,g_1} \cap G_{w_2,g_2}| \leq 1 \notag 
\end{eqnarray}

\subsubsection{SCE: Set Constraint Encoding}

From the set constraint model proposed previously, our $\Leftrightarrow_{enc}$ encoding rule automatically generates SAT instances as describe in Section \ref{sec:encoding}.
For each type of the above constraints we give the number of  clauses generated in the SAT instance:  

\paragraph{$p$ players per group every weeks:}
Constraints~(\ref{s1}) generates 
\[w.g.w.(g.p+ \sum_{i=1}^{g.p}
[
2u_i^{g.p}
(\lfloor \frac{u_i^{g.p}}{2}\rfloor+1)(\lceil \frac{u_i^{g.p}}{2}\rceil+1) 
-(\frac{u_i^{g.p}}{2}+1)
]) 
\] 
clauses
with $u_{g.p}^{g.p}=1$,$u_1^{g.p}={g.p}$ and  $u_i^{g.p}=u_{2i-1}^{g.p}+2u_{2i}^{g.p}+u_{2i+1}^{g.p}$. The complexity of the formula generated by Constraints~(\ref{s1}) is ${\cal O}(w^2.g^3.p^2)$.

\paragraph{Every golfer plays every week:}
Constraints~(\ref{s3}) generates $w.g.p$ clauses.

\paragraph{Two players cannot play twice together in the same group:}
Two formulations are possible:
\begin{itemize}
\item with implication formulation, Constraints~(\ref{s4}) generates $w.(w-1).g.(g+1).q.(q-1)/2)$ clauses (${\cal O}(w^2.g^4.p^2)$).
\item with cardinality formulation,  Constraints~(\ref{s5}) generates $w.((w-1)/2).g.((g+1)/2).3.q.(q+ \sum_{i=1}^{q}
[
2u_i^{q}
(\lfloor \frac{u_i^{q}}{2}\rfloor+1)(\lceil \frac{u_i^{q}}{2}\rceil+1) 
-(\frac{u_i^{q}}{2}+1)
]) $ clauses (${\cal O}(w^2.g^5.p^3)$).
\end{itemize}

\paragraph{Complexity of the generated SAT instances}
Complexity of Constraints~(\ref{s4}) is ${\cal O}(w^2.g^4.p^2)$ whereas complexity of Constraints~(\ref{s5}) is  ${\cal O}(w^2.g^5.p^3)$. Thus in the following we will only focus on the implication formulation (Constraints~(\ref{s4})). To summarize, the complexity of the SAT instances generated by the SCE model (Set Constraint Encoding model) made from Constraints~(\ref{s1}), (\ref{s3}), and (\ref{s4}) is  ${\cal O}(w^2.g^4.p^2)$.
In Section~\ref{modelstructure}, we show the exact numbers of clauses that are required for specific instances of the Social Golfer Problem. 

\paragraph{Post-treatment by Unit Propagation}
\label{sec:up}
Unit propagation is a simply process corresponding to constraint propagation. The idea is to eliminate unit clauses (clauses with only false literals and one free literal) by valuing the free literal to $true$. This valuation can produce new unit clauses and then the process is achieved until there is no longer any unit clause. In term of complexity, algorithms for unit propagation is in polynomial time; however, in practice, this process is insignificant compared to solving time and may significantly reduce:
\begin{itemize}
\item instances size,
\item number of variables,
\item  and solving time.
\end{itemize}
Note also that the cardinality constraint encoding that we have chosen generates a lot of unit clauses that vanish using unit propagation.

\section{Symmetry Breaking for the Social Golfer Problem} 
\label{sb}
The idea of symmetry breaking is to remove uninteresting solutions and to ease the work of a (SAT) solver. The Social Golfer Problem is highly symmetric: the position of a player in a group is not relevant; the  groups in a week can be renumbered;  the weeks can be swapped. Symmetry breaking thus consists in eliminating these symmetries by adding new constraints or modifying the model. \cite{GentLynce2005} proposes some clauses to remove symmetries among players, to order groups within a week with respect to their first player, to order lexicographically the weeks with respect to the second player in the first group of each week, ... However, these clauses become more and more complicated and mistakes can easily be introduced. Indeed, \cite{TriskaMusliu2012} revised the clauses for symmetry breaking of \cite{GentLynce2005} in order to correct the ranges of the various $\bigvee$ and $\bigwedge$ appearing in these clauses.

More symmetries can be broken, such as in \cite{frisch2002} or \cite{flener2002}. All symmetries can be broken, such as shown in \cite{crawford1996}, but this is often at the cost of a super exponential number of constraints. Thus, this cannot be considered in practice.

\subsection{Symmetry Breaking for TME}
\label{sbp}

In  \cite{TriskaMusliu2012}, three types of symmetry breaking are added to the TME encoding. Note that this is done by adding constraints. The first one consists in breaking the symmetry among players within each group.
\begin{eqnarray}
 \bigwedge_{i=1}^{x}
\bigwedge_{j=1}^{p-1}
\bigwedge_{k=1}^{g}
\bigwedge_{l=1}^{w}
\bigwedge_{m=1}^{i}
\neg G_{ijkl} \vee \neg G_{m(j+1)kl}
\label{sbtme1}
\end{eqnarray}

The second one consists in ordering all groups within a single week by their first players.
\begin{eqnarray}
 \bigwedge_{i=1}^{x}
\bigwedge_{k=1}^{g-1}
\bigwedge_{l=1}^{w}
\bigwedge_{m=1}^{i-1}
\neg G_{i1kl} \vee \neg G_{m1(k+1)l}
\label{sbtme2}
\end{eqnarray}

The last one consists in  strictly ascending second players in the first group of each week.
\begin{eqnarray}
 \bigwedge_{i=1}^{x}
\bigwedge_{l=1}^{w}
\bigwedge_{m=1}^{i}
\neg G_{i21l} \vee \neg G_{m21(l+1)}
 \label{sbtme3}
\end{eqnarray}

The encoding TME$^{\textrm{SB}}$ corresponding to the Triska-Musliu encoding with the above symmetry breaking is thus defined by Formulae~(\ref{atleastonceperweek})--(\ref{atmostoneplayerperposition}), (\ref{tmvar}), (\ref{tmc}), (\ref{sbtme1})--(\ref{sbtme3}).

\subsection{Symmetry Breaking with Set Constraint Model}
\label{sbp}

With our set constraint language, we have two possibilities to break symmetries. The first one consists in adding some constraints to the initial model; the second one consists in refining the model itself by modifying the supports of sets and the constraints.

Since our model is different from the one of~\cite{GentLynce2005,TriskaMusliu2012}, we do not obtain the same symmetries. However, we try to break similar symmetries as in~\cite{GentLynce2005,TriskaMusliu2012}.

The first group of symmetry breaking ($SB1$) consists in filling the first week as follows: the first $p$ players are sent to the first group of the first week; the   next $p$ players, on the second group of the first week; and so on.

We consider a second group $SB2$ of symmetry breaking which completes $SB1$. $SB2$ consists in spreading the first $p$ players (who already played together the first week in the first group due to $SB1$) in different groups each week: the first player in the first group of each week (except the first week); the second one in the second group of each week; and so on. This approximately corresponds to group $(23)$ of constraints of~\cite{TriskaMusliu2012}. 

We first consider the following fact to simplify the following models: when $p$ (the number of players per group) becomes greater than $g$ (the number of groups per week) we can rather obviously see that the problem has no solution. Indeed, consider the $p$ players of the first group of the first week; for the second week, they all must play in different groups; thus, the number of groups needs to be greater or equal to the number of players per group, otherwise, there is no solution. In the following, we thus consider $g\geq p$. However, if one does not want to make this simplification, it is sufficient to change $p$ by $min(g,p)$ in the following, and to add the constraints "Two players cannot play twice together in the same group" between $G_{1,1}$ and the other groups. Indeed, these constraints make immediately the model unsatisfiable for $g<p$.

\subsubsection{Symmetry breaking for the set constraint model by adding constraints}

In this section constraints are added to the initial model in order to break symmetries.
For $SB1$, we only have to add the following simple constraints to the model of the $SCE$.
\begin{eqnarray}
\forall i \in [1..p.g], p_i \in G_{1, i ~div~ (p + 1)} \label{sb1c}
\end{eqnarray}

For the second group $SB2$ of symmetry breaking, the required constraints are also simple:
\begin{eqnarray}
\forall i \in [2..w], \forall j \in [1..p], p_j \in G_{i,j}\label{sb2c}
\end{eqnarray}

We can note that these constraints add clauses to the set model and its SAT encoding, but all these extra constraints are unit clauses that will produce unit propagation and thus they will vanish.

The SAT encoding of the set model with symmetry breaking by adding constraints to the model is named SCE$^{\textrm{SBC}}$ and consists in Constraints~(\ref{s1}), (\ref{s3}), (\ref{s4}), (\ref{sb1c}), and~(\ref{sb2c}).

\subsubsection{Symmetry breaking for the set constraint model by modifying the model}

Modifying the model is more tedious. However, the gain is to reduce the supports of sets and cardinality constraints. These modified models will thus significantly reduce the size of the generated SAT instances.

The only modification for $SB1$  consists in both modifying the supports of the groups of the first week and to fix these groups: 
\[
\fa i \in [1..g], {\cal G}_{1,i}=\{p_{1+(i-1).g}, \ldots p_{p+(i-1).g} \}
\]
and
\begin{eqnarray}
\fa i \in [1..g], G_{1,i}={\cal G}_{1,i} \label{sb1m}
\end{eqnarray}
The other sets, variables, and constraints remain unchanged.

To introduce $SB2$, we change the group variables. Instead of the $G_{i,j}$, we now consider the sets $G'_{1,1}, \ldots, G'_{w,g}$ such that:
\begin{itemize}
\item for the first week $G_{i,j}=G'_{i,j}$;
\item for the following weeks $G_{i,j}=G'_{i,j} \cup \{p_j\}$ if $j \leq p$, $G_{i,j}=G'_{i,j}$ otherwise.
\end{itemize}  

The support of the $G'_{1,i}$ (i.e., the groups of the first week) are defined as with $SB1$. Since the $min(p,g)$ first player are spread on the  $min(p,g)$ first groups of each week, the supports of the other groups can be reduced. Let ${\cal P'}=\{ p_{min(p,g)+1},  \ldots , p_{q} \}$ be the set of golfers except the first ones. The supports can thus be defined by:
\[
\fa i \in [2..w], \forall j \in [1..g], {\cal G}_{i,j}={\cal P'}
\]

Constraints are modified as follows.

\paragraph{$P$ players per group every weeks:}
Constraints~(\ref{s1}) must be replaced by Constraints~(\ref{sb2m1})--(\ref{sb2m3}).
\begin{eqnarray}
\forall i \in [1..g], ~~ |G'_{1,i}|=p  \label{sb2m1}\\
\forall j \in [2..w], \forall i \in [1.. p], ~~ |G'_{j,i}|=p- 1  \label{sb2m2}\\
\forall j \in [2..w], \forall i \in [p+1..g], ~~ |G'_{j,i}|=p \label{sb2m3}
\end{eqnarray}

\paragraph{Every golfer plays every week:} Constraints~(\ref{sb2m4}) replace Constraints~(\ref{s3}).
\begin{eqnarray}
\forall j \in [2..w]~  \bigcup_{i=1..g} G_{j,i} = {\cal P'} \label{sb2m4}
\end{eqnarray}

\paragraph{Two players cannot play twice together in the same group:}

Constraints~(\ref{s4}) are replaced by Constraints~(\ref{sb2m5})--(\ref{sb2m8}).\\

We recall here that we are working on $G'_{i,j}$ which has the following relation with the intial set $G_{i,j}$ of the model without symmetry breaking: if  $j \leq p$ and $i>1$, then $G_{i,j}=G'_{i,j} \cup \{p_j\}$. Since 2 groups  $G_{i,j}$ with  $j \leq p$ and $i>1$ have player $p_j$ in common, the corresponding groups $G'_{i,j}$ (which supports do not contain the $p_l$, $l \leq p$) cannot have any other player $p_k$ in common:
\begin{eqnarray}
\begin{array}{r}
\forall w_1, w_2 \in [2..w], ~
 p_i \in {\cal P}, ~
 g_1 \in [1..p], ~
w_1 > w_2,
\\
        p_i \in G'_{w_1,g_1} 
    ~\rightarrow~ 
    p_i \not \in G'_{w_2,g_1}
\end{array}\label{sb2m5}
\end{eqnarray}
The relation between other two groups is not changed as shown below.\\

Constraints between a group of the first week (except the first group) and groups of other weeks:
\begin{eqnarray}
\begin{array}{r}
\forall w_1 \in [2..w], ~
 p_i, p_j \in {\cal P}, ~
 g_1 \in [2..g],  g_2 \in [1..g],~
 i > j ,\\
    p_i \in G'_{1,g_1} ~\wedge~ p_j \in G'_{1,g_1} ~\wedge~ p_i \in G'_{w_1,g_2}
    ~\rightarrow~ 
    p_j \not \in G'_{w_1,g_2}
\end{array}\label{sb2m6}
\end{eqnarray}
Note that if one does not consider the simplification $p\leq g$, then $g_1$ must be considered in $[2..g]$ to generate the proper constraints (that will generate a failure during the resolution of the SAT instance).\\

Constraints between two groups (except of the first week) equally numbered with an index greater than $p$:
\begin{eqnarray}
\begin{array}{r}
\forall w_1, w_2 \in [2..w], ~
 p_i, p_j \in {\cal P}, ~
 g_1 \in [p+1..g], ~
w_1 > w_2,~
 i > j ,\\
    p_i \in G'_{w_1,g_1} ~\wedge~ p_j \in G'_{w_1,g_1} ~\wedge~ p_i \in G'_{w_2,g_1}
    ~\rightarrow~ 
    p_j \not \in G'_{w_2,g_1}
\end{array}\label{sb2m7}
\end{eqnarray}

Constraints between two groups (except of the first week) not equally numbered :
\begin{eqnarray}
\begin{array}{r}
\forall w_1, w_2 \in [2..w], ~
 p_i, p_j \in {\cal P}, ~
 g_1, g_2 \in [1..g], ~
w_1 > w_2, ~
g_1 \not = g_2,~
 i > j ,\\
    p_i \in G'_{w_1,g_1} ~\wedge~ p_j \in G'_{w_1,g_1} ~\wedge~ p_i \in G'_{w_2,g_2}
    ~\rightarrow~ 
    p_j \not \in G'_{w_2,g_2}
\end{array}\label{sb2m8}
\end{eqnarray}

The SAT encoding of the set model with symmetry breaking by modifying the model is named SCE$^{\textrm{SBM}}$ and consists in Constraints~(\ref{sb1m})--(\ref{sb2m8}).

\section{Comparisons of Models}
\label{sec:compmod}
Table~\ref{tab:desc}  summarizes the various encodings that we will compare in the following sections. These encodings have been described in previous sections. NAME$_{\textrm{UP}}$ denotes the  encoding NAME after unit propagation.

\begin{table*}[!hbt]
\begin{center}
\begin{normalsize}
\caption{List of the encoding names, descriptions and the corresponding constraints or formulae.}
\label{tab:desc}
\begin{tabular}{|l|l|c|}
\hline
Encoding &Description&Corresponding constraints\\
 Name&&or formulae\\
\hline
\hline
DE&Direct Encoding&(\ref{atleastonceperweek})--(\ref{social})\\
\hline
TME&Triska-Musliu encoding&(\ref{atleastonceperweek})--(\ref{atmostoneplayerperposition}), (\ref{tmvar}), (\ref{tmc})\\
\hline
TME$^{\textrm{SB}}$&TME with symmetry breaking&(\ref{atleastonceperweek})--(\ref{atmostoneplayerperposition}), (\ref{tmvar}), (\ref{tmc}), (\ref{sbtme1})--(\ref{sbtme3})\\
\hline
SCE&SAT encoding of the set&(\ref{s1}), (\ref{s3}), (\ref{s4})\\
&constraint model&\\
\hline
SCE$^{\textrm{SBC}}$&SCE with with symmetry&(\ref{s1}), (\ref{s3}), (\ref{s4}), (\ref{sb1c}), (\ref{sb2c})\\
&breaking by adding constraints&\\
\hline
SCE$^{\textrm{SBM}}$&SCE with with symmetry&(\ref{sb1m})--(\ref{sb2m8})\\
&breaking by modifying the model&\\
\hline
NAME$_{\textrm{UP}}$&encoding after unit propagation &\\
&treatment&\\
\hline
\end{tabular}
\end{normalsize}
\end{center}
\end{table*}

\subsection{Declarativity}

We compare here the models in terms of declarativity. Comparisons in terms of structures (number of clauses, number of variables) are given in the next section.

The first remark is that the variables we use in the set model are much simpler. Indeed, we have only two indices instead of 4, making them more readable. This is due to the fact that we do not have to number the positions in a group (groups are sets), and we do not have to add an index for the number of  players (players are members of the groups).


The second difference to be noticed is the simplicity and declarativity of constraints. Indeed, set constraints are more declarative than pure SAT  clauses. Then, the encoding in SAT is performed using the encoding rules $\Leftrightarrow_{enc}$. The advantage is double:
\begin{itemize}
\item first, constraints are readable, declarative, easy to modify, resulting in a much understandable model;
\item second, less mistakes are introduced since the modeling process is much simpler.
\end{itemize}

Last, but not least, the set encoding is solver independent. Indeed, the same model (changing the syntax) could be used in a CSP solver with set constraints or in a SAT solver after applying the rule encoding $\Leftrightarrow_{enc}$ proposed above.

Adding symmetry breaking in the direct encodings DE and TME can only be done by adding constraints/clauses. With the set model, symmetry breaking can also be done by modifying the model itself. The process is a bit more complicateed than just adding constraints, but the result is worth: instances are smaller and solving time is faster.

To summarize, in terms of declarativity, readability, error introduction, and solver dependence, our set model is superior to  direct encodings such as DE or TME. Breaking symmetries is also easier in the set model.

Each encoding produces specific SAT instances. We compare the direct encodings and the set constraint encoding in two ways: the size of the provided instances and the ease to solve them with a complete SAT solver.

\subsection{Model Structure}
\label{modelstructure}
In order to compare our set constraint encoding, we generate a set of social golfer instances with: the direct encoding DE, the Triska-Musliu encoding (TME) proposed in \cite{TriskaMusliu2012}, and our set constraint encoding with unit propagation post-treatment (SCE$_{\textrm{UP}}$) and without (SCE). 
In Table \ref{tab:instances}, each instance is defined by the triple (groups, players per group, weeks) and for each encoding the number of variables and the number of (generated) clauses are provided. It is not possible to compare efficiency of an encoding only in terms of instance size (this is done in the next section). Nevertheless, big instances are intractable due to the limited size of computer memory. It is thus necessary to generate instances as small as possible. In Table \ref{tab:instances}, for each instance, encodings generating the smallest number of clauses and variables are  in bold.

\renewcommand{\tabcolsep}{0.15cm}
\begin{table}
\caption{Size of instances generated using the direct encoding (DE), the Triska and Musliu encoding (TME) \cite{TriskaMusliu2012}, the set constraints encoding (with unit propagation post-process (SCE$_{\textrm{UP}}$) and without (SCE)).}
\label{tab:instances}
\begin{scriptsize}
\begin{center}
\begin{tabular}{|l||r|r||r|r||r|r||r|r|}
\hline
\multicolumn{1}{|c||}{Prob.}&\multicolumn{2}{c||}{DE}&\multicolumn{2}{c||}{TME}&\multicolumn{2}{c||}{SCE}&\multicolumn{2}{c|}{SCE$_{\textrm{UP}}$}\\
\multicolumn{1}{|c||}{}&\#Vars&\#Cls&\#Vars&\#Cls&\#Vars&\#Cls&\#Vars&\#Cls\\
\hline
5-3-6	&1 350	&3 203 055	&1 800	&60 255	&8 625	&50 400	&1 410	&43 905	\\
5-3-7	&1 575	&4 481 085	&2 100	&79 485	&11 110	&67 985	&1 645	&60 410	\\
8-4-4	&4 096	&48 850 176	&5 120	&322 816	&24 224	&234 912	&\textbf{3 840}	&\textbf{204 928}	\\
8-4-5	&5 120	&81 378 880	&6 400	&482 880	&34 752	&372 992	&\textbf{4 800}	&\textbf{335 520}	\\
8-4-6	&6 144	&121 896 960	&7 680	&674 688	&47 072	&542 816	&\textbf{5 760}	&\textbf{497 856	}\\
8-4-7	&7 168	&170 815 680	&8 960	&898 240	&61 184	&744 384	&\textbf{6 720}	&\textbf{691 936}	\\
8-4-8	&8 192	&227 723 776	&10 240	&1 153 536	&77 088	&977 696	&\textbf{7 680}	&\textbf{917 760}	\\
8-4-9	&9 216	&292 552 704	&11 520	&1 440 576	&94 784	&1 242 752	&\textbf{8 640}	&\textbf{1 175 328}	\\
8-4-10	&10 240	&365 690 880	&12 800	&1 759 360	&114 272	&1 539 552	&\textbf{9 600}	&\textbf{1 464 640}	\\
9-4-6	&7 776	&196 150 032	&9 720	&1 047 762	&117 324	&858 366	&\textbf{7 344}	&\textbf{792 882	}\\
9-4-7	&9 072	&274 564 584	&11 340	&1 400 994	&157 284	&1 180 026	&\textbf{8 568}	&\textbf{1 103 634}	\\
9-4-8	&10 368	&366 042 816	&12 960	&1 805 256	&203 076	&1 552 716	&\textbf{9 792}	&\textbf{1 465 416}	\\
9-4-9	&11 664	&470 584 728	&14 580	&2 260 548	&254 700	&1 976 436	&\textbf{11 016}	&\textbf{1 878 228}	\\
9-4-10	&12 960	&588 190 320	&16 200	&2 766 870	&312 156	&2 451 186	&\textbf{12 240}	&\textbf{2 342 070}	\\
\hline
\end{tabular}
\end{center}
\end{scriptsize}
\end{table}

Direct encoding (DE) is clearly unsuitable when the number of players or groups increases: the number of clauses immediately blows up. With the introduction of auxiliary variables the number of clauses is less important for TME but the number of variables is increased. SCE produces more variables but less clauses. As might be expected, SCE$_{\textrm{UP}}$ provides the most interesting encoding in terms of number of clauses and number of variables: indeed, SCE generates a lot of unit clauses and binary clauses (Section \ref{sec:directmodel}) than vanish using unit propagation.

\subsection{Impact of the symmetry breaking}
Social Golfer Problem has a lot of identical solutions modulo symmetries. In Table \ref{tab:sym} we apply the two symmetry breaking processes presented in Section \ref{sbp}  to the instances proposed in Table \ref{tab:instances}. 

For TME, introducing symmetry breaking constraints only increases the number of clauses (around 10\% more clauses), the number of variables does not change. Note also that unit propagation is not worth for TME instances nor for TME$^{\textrm{SB}}$ instances: there is no unit clause and the size of the instance is not changed (both in terms of variables and clauses).

For SCE, symmetry breaking by adding constraints adds a negligible amount of constraints (see SCE$^{\textrm{SBC}}$). More interestingly, adding symmetry breaking by modifying the model (SCE$^{\textrm{SBM}}$) significantly reduces the size of the generated SAT instances: from 20 up to 60\% less variables and from 40 to 60\% less clauses. This significant reduction is due to the reduction of supports and to the cardinality constraints: sets with $k-1$ elements instead of $k$, and less clauses are necessary when supports are smaller.

Without unit propagation, the instances of SCE$^{\textrm{SBM}}$ are always the smallest one generated with respect to the number of clauses.

\medskip

Unit propagation has no impact at all on TME. However, its impact is significant on SCE, SCE$^{\textrm{SBM}}$, and SCE$^{\textrm{SBC}}$:
\begin{itemize}
\item for SCE, unit propagation divides the number of variables by 6 to 25: this is mainly due to the variables of the cardinality constraints. The number of clauses is reduced of around 10\%.
\item for SCE$^{\textrm{SBC}}$, unit propagation reduce even more the number of variables (up to 30 times less variables). The number of clauses is reduced from 30 to 60\%.
\item for SCE$^{\textrm{SBM}}$, unit propagation is less spectacular: indeed, the initial model itself is reduced by adding symmetry breaking. However, the number of variable is divided by 5 up to 15. The number of clauses is reduced of about 10\%.

\end{itemize}
To summarize, unit propagation is more beneficial to SCE$^{\textrm{SBC}}$; however, SCE$_{\textrm{UP}}^{\textrm{SBM}}$ always gives the best instances in terms of number of clauses and number of variables.

\begin{table}[hbt!]
\caption{Size of instances generated using Triska-Musliu encoding and the set constraint encoding with symmetry breaking (TME$^{\textrm{SB}}$ Triska and Musliu encoding with symmetry breaking, SCE$^{\textrm{SBM}}$ for symmetry breaking in the supports and SCE$^{\textrm{SBC}}$ for symmetry breaking by adding constraints).}
\label{tab:sym}
\begin{center}
\begin{tabular}{|l||r|r||r|r||r|r|}
\hline
Prob.&\multicolumn{3}{c||}{TME}&\multicolumn{3}{c|}{TME$^{\textrm{SB}}$ and TME$_{\textrm{UP}}^{\textrm{SB}}$}\\
	&var	&\multicolumn{2}{c||}{clauses} &var	&\multicolumn{2}{c|}{clauses}\\
\hline
5-3-6	&1 800	&\multicolumn{2}{c||}{60 255	}	&1 800	&\multicolumn{2}{c|}{70 935}\\
5-3-7	&2 100	&\multicolumn{2}{c||}{79 485	}	&2 100	&\multicolumn{2}{c|}{91 965	}\\
8-4-4	&5 120	&\multicolumn{2}{c||}{322 816}	&5 120	&\multicolumn{2}{c|}{389 872}\\
8-4-5	&6 400	&\multicolumn{2}{c||}{482 880}	&6 400	&\multicolumn{2}{c|}{566 832}\\
8-4-6	&7 680	&\multicolumn{2}{c||}{674 688}	&7 680	&\multicolumn{2}{c|}{775 536}\\
8-4-7	&8 960	&\multicolumn{2}{c||}{898 240}	&8 960	&\multicolumn{2}{c|}{1 015 984}\\
8-4-8	&10 240	&\multicolumn{2}{c||}{1 153 536}	&10 240	&\multicolumn{2}{c|}{1 288 176}\\
8-4-9	&11 520	&\multicolumn{2}{c||}{1 440 576}	&11 520	&\multicolumn{2}{c|}{1 592 112}\\
8-4-10	&12 800	&\multicolumn{2}{c||}{1 759 360	}	&12 800	&\multicolumn{2}{c|}{1 927 792}\\
9-4-6	&9 720	&\multicolumn{2}{c||}{1 047 762	}	&9 720	&\multicolumn{2}{c|}{1 190 952}\\
9-4-7	&11 340	&\multicolumn{2}{c||}{1 400 994	}	&11 340	&\multicolumn{2}{c|}{1 568 160}\\
9-4-8	&12 960	&\multicolumn{2}{c||}{1 805 256	}	&12 960	&\multicolumn{2}{c|}{1 996 398}\\
9-4-9	&14 580	&\multicolumn{2}{c||}{2 260 548	}	&14 580	&\multicolumn{2}{c|}{2 260 548}\\
9-4-10	&16 200	&\multicolumn{2}{c||}{2 766 870	}	&16 200	&\multicolumn{2}{c|}{3 005 964}\\
\hline
\hline
Prob.	&\multicolumn{2}{c||}{SCE}&\multicolumn{2}{c||}{SCE$^{\textrm{SBM}}$}&\multicolumn{2}{c|}{SCE$^{\textrm{SBC}}$}\\
	&\multicolumn{1}{c|}{var}	&\multicolumn{1}{c||}{clauses}&\multicolumn{1}{c|}{var}	&\multicolumn{1}{c||}{clauses}&\multicolumn{1}{c|}{var}	&\multicolumn{1}{c|}{clauses}\\
\hline
5-3-6	&8 625	&50 400	&5 702	&21 487	&8 625	&50 430\\
5-3-7	&11 110	&67 985	&7 734	&30 243	&11 110	&68 018\\
8-4-4	&24 224	&234 912	&14 192	&95 712	&24 224	&234 956\\
8-4-5	&34 752	&372 992	&22 476	&173 180	&34 752	&373 040\\
8-4-6	&47 072	&542 816	&32 552	&273 440	&47 072	&542 868\\
8-4-7	&61 184	&744 384	&44 420	&396 492	&61 184	&744 440\\
8-4-8	&77 088	&977 696	&58 080	&542 336	&77 088	&977 756\\
8-4-9	&94 784	&1 242 752	&73 532	&710 972	&94 784	&1 242 816\\
8-4-10	&114 272	&1 539 552	&90 776	&902 400	&114 272	&1 539 620\\
9-4-6	&117 324	&858 366	&46 344	&447 832	&117 324	&858 422\\
9-4-7	&157 284	&1 180 026	&63 368	&652 344	&157 284	&1 180 086\\
9-4-8	&203 076	&1 552 716	&82 984	&895 176	&203 076	&1 552 780\\
9-4-9	&254 700	&1 976 436	&105 192	&1 176 328	&254 700	&1 976 504\\
9-4-10	&312 156	&2 451 186	&129 992	&1 495 800	&312 156	&2 451 258\\
\hline
\hline
Prob.	&\multicolumn{2}{c||}{SCE$_{\textrm{UP}}$}&\multicolumn{2}{c||}{SCE$_{\textrm{UP}}^{\textrm{SBM}}$}&\multicolumn{2}{c|}{SCE$_{\textrm{UP}}^{\textrm{SBC}}$}\\
	&\multicolumn{1}{c|}{var}	&\multicolumn{1}{c||}{clauses}&\multicolumn{1}{c|}{var}	&\multicolumn{1}{c||}{clauses}&\multicolumn{1}{c|}{var}	&\multicolumn{1}{c|}{clauses}\\
\hline
5-3-6	&1 410	&43 905	&\textbf{860}	&\textbf{17 680}	&980	&23 110\\
5-3-7	&1 645	&60 410	&\textbf{1 032}	&\textbf{25 680}	&1 176	&33 690\\
8-4-4	&3 840	&204 928	&\textbf{2 376}	&\textbf{77 700}	&2 580	&91 548\\
8-4-5	&4 800	&335 520	&\textbf{3 168}	&\textbf{149 184}	&3 440	&176 240\\
8-4-6	&5 760	&497 856	&\textbf{3 960}	&\textbf{243 460}	&4 300	&288 020\\
8-4-7	&6 720	&691 936	&\textbf{4 752}	&\textbf{360 528	}&5 160	&426 888\\
8-4-8	&7 680	&917 760	&\textbf{5 544}	&\textbf{500 388}	&6 020	&592 844\\
8-4-9	&8 640	&1 175 328	&\textbf{6 336}	&\textbf{663 040}	&6 880	&785 888\\
8-4-10	&9 600	&1 464 640	&\textbf{7 128}	&\textbf{848 484}	&7 740	&1 006 020\\
9-4-6	&7 344	&792 882	&\textbf{5 620}	&\textbf{471 690}	&5 620	&471 690\\
9-4-7	&8 568	&1 103 634	&\textbf{6 008}	&\textbf{561 712}	&6 744	&700 830\\
9-4-8	&9 792	&1 465 416	&\textbf{7 024}	&\textbf{782 620}	&7 868	&974 904\\
9-4-9	&11 016	&1 878 228	&\textbf{8 040}	&\textbf{1 039 956}	&8 992	&1 293 912\\
9-4-10	&12 240	&2 342 070	&\textbf{9 056}	&\textbf{1 333 720}	&10 116	&1 657 854\\
\hline
\end{tabular}
\end{center}
\end{table}

\section{Experimental Analysis}
\label{sec:analysis}
\begin{table}[hbt!]
\caption{Minisat \textbf{with} SatElite: Running time for the set constraints encoding and the Triska-Musliu encoding. Formulations with symmetry breaking and unit propagation are compared.}
\label{tab:time1}
\begin{center}
\begin{tabular}{|l||r|r|r|r|r|r|r|r|}
\hline
Prob.	&TME	&TME$^{\textrm{SB}}$	&SCE	&SCE$^{\textrm{SBM}}$	&SCE$^{\textrm{SBC}}$&SCE$_{\textrm{UP}}$	&SCE$_{\textrm{UP}}^{\textrm{SBM}}$	&SCE$_{\textrm{UP}}^{\textrm{SBC}}$\\
\cline{2-9}
&\multicolumn{8}{c|}{Time in seconds (limited to 300)}\\
\hline
5-3-6	&8.92	&0.69	&0.18	&0.06	&0.12	&0.12	&0.07	&\textbf{0.04}\\
5-3-7	&98.28	&13.37	&1.42	&0.13	&1.21	&5.09	&0.09	&\textbf{0.08}\\
8-4-4	&1.04	&1.33	&0.97	&0.32	&1.19	&0.90	&0.29	&\textbf{0.27}\\
8-4-5	&2.26	&2.64	&1.93	&0.86	&2.51	&1.89	&0.84	&\textbf{0.78}\\
8-4-6	&4.44	&5.16	&3.65	&1.87	&4.74	&3.65	&1.82	&\textbf{1.71}\\
8-4-7	&34.25	&94.68	&8.66	&3.59	&8.52	&7.52	&3.64	&\textbf{3.46}\\
8-4-8	&-	&-	&-	&-	&-	&-	&-	&-\\
8-4-9	&-	&-	&-	&-	&-	&-	&-	&-\\
8-4-10	&-	&-	&-	&-	&-	&-	&-	&-\\
9-4-6	&8.45	&10.52	&11.24	&3.15	&10.34	&11.10	&\textbf{2.71}	&4.58\\
9-4-7	&13.69	&27.16	&18.95	&5.80	&17.8	&19.04	&\textbf{5.12}	&8.76\\
9-4-8	&-	&-	&31.87	&\textbf{11.10}	&29.60	&31.48	&12.72	&14.90\\
9-4-9	&-	&-	&-	&-	&-	&-	&-	&-\\
9-4-10	&-	&-	&-	&-	&-	&-	&-	&-\\
\hline
\end{tabular}
\end{center}
\end{table}

\begin{table}[!hbt]
\caption{Minisat \textbf{without} SatElite: Running time for the set constraints encoding and the Triska-Musliu encoding. Formulations with symmetry breaking and unit propagation are compared.}
\label{tab:time2}
\begin{center}
\begin{tabular}{|l||r|r|r|r|r|r|r|r|}
\hline
Prob.	&TME	&TME$^{\textrm{SB}}$	&SCE	&SCE$^{\textrm{SBM}}$	&SCE$^{\textrm{SBC}}$&SCE$_{\textrm{UP}}$	&SCE$_{\textrm{UP}}^{\textrm{SBM}}$	&SCE$_{\textrm{UP}}^{\textrm{SBC}}$\\
\cline{2-9}
&\multicolumn{8}{c|}{Time in seconds (limited to 300)}\\
\hline
5-3-6	&9.37	&0.30	&1.05	&\textbf{0.01}	&\textbf{0.01}	&0.26	&\textbf{0.01}	&\textbf{0.01}\\
5-3-7	&97.47	&24.86	&9.19	&\textbf{0.06}	&0.13	&5.67	&1.79	&0.28\\
8-4-4	&0.05	&0.23	&0.09	&\textbf{0.03}	&0.07	&0.07	&\textbf{0.03}	&\textbf{0.03}\\
8-4-5	&0.08	&0.58	&0.13	&0.06	&0.11	&0.06	&\textbf{0.05}	&0.07\\
8-4-6	&0.25	&3.58	&0.27	&0.14	&0.18	&0.19	&\textbf{0.08}	&0.09\\
8-4-7	&27.05	&25.88	&3.53	&\textbf{0.48}	&1.71	&1.94	&0.56	&0.98\\
8-4-8	&-	&-	&-	&-	&-	&-	&-	&-\\
8-4-9	&-	&-	&-	&-	&-	&-	&-	&-\\
8-4-10	&-	&-	&-	&-	&-	&-	&-	&-\\
9-4-6	&0.23	&3.72	&0.37	&0.13	&0.29	&0.25	&\textbf{0.11}	&0.13\\
9-4-7	&0.31	&6.61	&0.58	&0.22	&0.51	&0.36	&\textbf{0.14}	&0.24\\
9-4-8	&247.83	&-	&14.66	&5.03	&1.10	&20.93	&2.62	&\textbf{0.68}\\
9-4-9	&-	&-	&-	&-	&-	&-	&-	&-\\
9-4-10	&-	&-	&-	&-	&-	&-	&-	&-\\
\hline
\end{tabular}
\end{center}
\end{table}

In the previous section we have shown that SCE enables us to obtain the smallest instances with unit propagation. The use of symmetry breaking also reduces the size of the SAT instances. It can happen that  symmetry breaking makes more difficult the resolution: by changing the search landscape, an "easy" solution can disappear; with incomplete solvers (such as local search), symmetry breaking can partitions the search space and makes difficult a path to a solution.  In this section we will compare the efficiency of the encodings in terms of running time.

To compare our set constraints encoding with Triska-Musliu \cite{TriskaMusliu2012} encoding, we use the well known solver Minisat \cite{minisat03}. This solver won various competitions~\footnote{http://www.satcompetition.org/}. Since some few years, a pre-treatment named SatELite~\cite{satelite05} has been added to Minisat in order to drastically reduce the number of clauses (e.g., by using subsumptions detections) and variables (e.g., eliminating pure literals). This pre-treatment has a cost in terms of running time but it generally improves the global running time. It is now included in Minisat but an option enables one to desactivate it.

Experimentations are realized on a 2.60GHz Intel Core i5-2540M CPU and 4 GB RAM. For each experiment, the time-out is $300$ seconds. Larger execution times were tested but no real differences were observed. Results for the direct encoding DE are not presented since, as supposed, no results are obtained in a reasonable time.

Table \ref{tab:time1} and Table  \ref{tab:time2} represent respectively the running time of Minisat with the use of SatElite as pre-treatment and without pre-treatment. 

First of all, the two tables show that the use of SatElite is difficult to predict: for some instances, it significantly improves the results whereas for others, it significantly degrades the results. On average,  it  does not improve the results and the best running times are obtained without pre-treatment.

Moreover, symmetry breaking modifying the model (SCE$^{\textrm{SBM}}$) provides the best results (or results very close to the best ones), with or without pre-treament.
The use of unit propagation seems to have a weak impact to the resolution time of SCE$^{\textrm{SBM}}$.

Adding constraints to break symmetries (SCE$^{\textrm{SBC}}$) does not produce improvement except when unit propagation is applied (SCE$_{\textrm{UP}}^{\textrm{SBC}}$). Indeed, SCE$_{\textrm{UP}}^{\textrm{SBC}}$ obtain results as good as SCE$^{\textrm{SBM}}$.

Breaking symmetries in TME is rather fluctuating: depending on the instances and depending on the use of SatELite, it significantly improves or degrades the results.

To summarize, the best results are obtained with our set constraint model, with  SCE$_{\textrm{UP}}^{\textrm{SBC}}$ when the pre-treatment is applied, or predominantly with  SCE$_{\textrm{UP}}^{\textrm{SBM}}$ when the pre-treatment is not applied. Finally, the best results are obtained without pre-treatment.

\section{Discussion}
\label{discussion}

\paragraph{Modeling}
Modeling a problem with set constraints and then automatically generating the corresponding SAT instances is much simpler than writing directly encodings such as DE or TME. Breaking symmetries is rather tedious in direct encodings, very easy by adding constraints in the set model, and rather easy by modifying the set constraint model.

Using a higher level formalism (such as our set constraint) is thus beneficial to the modeling phase: it simplifies the task, and avoid making errors (mainly errors in the numerous indices required by a direct encoding). The SAT encoding is then automatically done.

\paragraph{SAT Instances}
We have shown that the SAT instances that are automatically produced by our encoding rules are of good quality: 
\begin{itemize}
\item they always produce significantly less clauses (with or without symmetry breaking, and with or without unit propagation);
\item with unit propagation, they also generate less variables;
\item and finally, they are solved faster with Minisat, without "tuning parameters", with or without pre-treatment with SatElite.
\end{itemize} 

\paragraph{Symmetry breaking}
We have shown that breaking symmetries by adding constraint to the set model is very simple. Moreover, the generated SAT instances after unit-propagation are much smaller, and the solving time is also improved.  

Symmetry breaking by modifying the model is even more beneficial. However, the effort for modifying the model is more important than the effort for adding constraints. This extra work is very beneficial for the size of the generated SAT instances, but not so much worth for the solving time (it is depending on instances, and pre-treatment).
Thus, one has to make the trade-off between solving time and modeling time. The size of the generated instances can be the deciding factor: larger problems can be modeled and generated introducing symmetry breaking into the model as in  SCE$^{\textrm{SBM}}$.

\paragraph{Set constraints in constraint programming}
The declarativity of set constraints in constraint programming (such as in~\cite{ConjuntoILPS94} or in~\cite{CLPS}) is more or less the same as the one of our set constraints in terms of sets: that was our goal. However, our approach is different: in systems such as~\cite{ConjuntoILPS94} or~\cite{choco}, sets constraints are not the only constraints, but a special set solver has to be designed to solve these models. For example, the mechanism of~\cite{ConjuntoILPS94} consists in reducing the domain of the sets by working on lower and upper bounds of the sets and to combine this process with search. Note that the domain of a set is similar to our notion of support, and lower and upper bounds of sets are the smallest and largest elements of a set with respect to a given ordering. Our approach is different: we do not want to design a special solver, nor to tune an existing one for efficiently solving our SAT instances; we want to transform a high level model written with set constraints into a good quality (in terms of size and solving time) SAT instance that is efficiently solved by an existing multi-purpose SAT solver.

Note that in the future, we want to add a pre-process to reduce support sizes. Indeed, the size of the SAT instances depends on the size of the supports. For the Social Golfer Problem, supports are minimal: they cannot be reduced without loosing solutions. But for some other problems, supports can be reduced by a deduction process (withtout loosing solution), and thus, generated SAT instances can be reduced. Such a process could be similar to one application of the first phase of the mechanism of~\cite{ConjuntoILPS94} without search.

Note also that in~\cite{DBLP:conf/csclp/Azevedo06} some comparisons of set constraint solvers in constraint programming are given for the social golfer problem. Most of the results reported are obtained by giving special (dynamic) search heuristics or special solving mechanisms. The approach is thus very different from ours.

\section{Conclusion}
\label{conclusion}

We have presented a technique for encoding set constraints into SAT:  the modeling process is achieved using some very declarative set constraints which are then automatically transformed into SAT variables and clauses using our$\Leftrightarrow_{enc}$ encoding rules. This technique has been applied successfully to model en encode the Social Golfer Problem, and to study some symmetry breaking on this problem.

The advantages of our technique are the following:
\begin{itemize}
\item the modeling process is simple, declarative, and readable. Moreover, it is solver independent and independent from CSP or SAT;
\item the technique is less error-prone than hand-written SAT encodings; 
\item breaking symmetry can be achieved by just adding new constraints or by refining the model (this cannot be done with direct encodings such as DE or TME);
\item the SAT instances which are automatically generated are smaller than the ones of~\cite{TriskaMusliu2012} that are  hand-made written and improved; with unit propagation, our instances also contain less variables than the ones of~\cite{TriskaMusliu2012}; 
\item finally, with respect to solving time, our automatically generated instances of the Social Golfer Problem are solved faster with or without unit propagation, with or without constraint breaking, with or without SatElite (the pre-treatment mechanism of Minisat).
\end{itemize}
We have tested our technique to model and solve other problems (such as n-queen problem, Sudoku, WhoWithWhom, \ldots). Each time we obtained very readable and simple set models. The generated SAT instances also appeared to be well-suited for Minisat.

\bigskip

In the future, we plan to use our set constraints encoding for formalizing domain variables and sequences of elements. To this end,
we will need to add some new constraints and to complete our $\Leftrightarrow_{enc}$ encoding rule.  

We want to refine the notion of supports and reduce their sizes. As said before, this do not have any impact on a problem such as the Social Golfer Problem for which supports are already minimal. But for many problems (in which supports are not clear at the principle), it is important to reduce the size of the supports (using a pre-treatment) before generating the SAT instances.

Finally, we also plan to combine set constraints with arithmetic constraints, and we want to define the corresponding combining SAT encoding.

\bibliographystyle{spmpsci}      

\end{document}